# An AI-Powered Culturally Aware Chatbot for Stress Detection and Wellness Support among Pakistani University Students Using NLP and Machine Learning

**Muhammad Fahad Bashir, Muhammad Afzal** , *Department of Computer Science & Information Technology, Ghazi University Dera Ghazi Khan , Pakistan*
*Email:bashirfahad789@gmail.com, afzal.ano1010@gmail.com*

## ABSTRACT

With the existing digital mental health tools specifically developed for Western settings, Pakistani students are exposed to a uniquely compounded stress situation in their university that includes academic, financial, familial, and relational stressors, which have become a serious concern for academic and psychological development of students in Pakistani universities. This paper introduces a new, AI-driven and culturally sensitive stress detection and wellness support system that is tailored to the context of Pakistani university students. The system is based on a machine learning model called Random Forest which is trained using a validated student stress data set of 1100 responses on 20 features from psychological, physiological, academic, environmental and social aspects, with an accuracy of 89.09% and a macro F1-score of 0.89, in three stress severity levels. The classification outputs are passed on to an open-source large language model through OpenRouter API, where an appropriately crafted system prompt, culturally aware, gives the model a conversation about wellness, in English, Urdu and Roman Urdu. The second most predictive stress factor in this population identified by feature importance analysis was teacher-student relationship, which is a culturally important stress factor highlighting the need for region-aware mental health systems. Future research will involve primary data collection from students at various academic levels of Pakistani Universities with the validated DASS-21 instrument focusing on the students who are moving from FSc to undergraduate studies, which is a time of being psychologically vulnerable which is under-researched.

## 1. INTRODUCTION

Student stress has become a major concern in Pakistan's universities, along with the academic pressure, there are financial, social, and family pressures that are much different that are experienced by students in western countries in universities [1, 2, 3]. In developed countries, students usually have to deal with one or two major stressors, but in Pakistan, university students have to cope with all four stressors at the same time: fee pressures, family expectations, social obligations, and demands for academic performance [4]. Unresolved, it can lead to poor academic achievement, poor interpersonal relations, and even depression, psychological burnout, in extreme cases [5, 6, 7]. Studies with Pak University students have also found high rates of stress, anxiety, and depression among this group, indicating the need to provide easily accessible mental health support for them [8, 9].

The increasing availability of digital mental health resources and AI-powered chatbots, however, does not match the needs of university students in Pakistan. The majority of mental health chatbots are designed and trained using datasets from Western contexts, which are culturally out of line with social contexts in South Asia [10]. Most of these tools are English language based, which poses a major communication challenge when students think, feel and express distress in Urdu or Roman Urdu. Moreover, current tools do not address an essential cultural fact that in Pakistan students do not often talk about their emotional issues with their parents, teachers or peers because of social stigma and cultural norms that deter sharing of such problems in the open [11]. In some areas, tools are available but often expensive and not available in less well-equipped universities in smaller regions.

In response to these gaps, the present paper suggests a culturally aware and AI based chatbot system called Sukoon which is specifically designed for Pakistani university students. Student stress information is collected using a validated 20-feature survey instrument and the system uses a machine learning classification model to gauge each student's level of stress in several categories. A culturally sensitive conversational AI chatbot provides tailored wellness guidance that is accessible to the audience and in a style that aligns with Pakistani cultural values based on the classification.

This paper makes several contributions, as follows. First, we introduce a machine learning-based stress classification model trained by validated student stress data containing 1100 responses, which has an accuracy of 89.09% among three stress levels [12, 13]. Second, we design and implement a culturally aware AI chatbot named Sukoon that will also provide stress level appropriate wellness support sensitive to Pakistani social and cultural norms based on Stepped Care Model [14]. Thirdly, we perform feature importance analysis that gives culturally important results of the predictive variables for student stress in the Pakistani context of university education. Finally, we introduce an end-to-end hybrid system that takes into account the integration of ML classification with LLM-based dialogue, which is the first hybrid system for this under-served population.

## 2. RELATED WORK

### *2.1 ML-Based Stress Classification*

There are several studies that have investigated machine learning methods for identifying and categorizing student stress through survey data. Ovi et al. [15] introduced a machine learning based approach for student stress level classification using ensemble and showed that the context-aware method is superior to periodic survey methods, based on two sets of data collected from surveys. Likewise, Zahra et al. [16] used a Random Forest classifier to classify the stress level of 143 university students using the Indonesian version of DASS-21 questionnaire, with a classification accuracy of 87.93%, showing the possibility of stress classification using ML in a relatively small stress dataset derived from a questionnaire. None of these studies, however, takes into consideration any culture and geographical settings and hence it is not applicable for the population of the region like Pakistani university students. Moreover, neither system offers any follow-up wellness intervention or chatbot-based support following the detection of stress — something that the present study addresses.

### *2.2 Student Mental Health Context in Pakistan*

There are some observational studies which have established empirical evidence of mental health problems among Pakistani University students.A few observational studies have found empirical evidence for mental health problems among students of the University of Pakistan. To establish the correlation between overthinking and mental health among 150 students from various universities of Multan district, Qasim et al. [6] found that there was a significant positive association among ruminating and worsening mental health outcomes. However, using standardised assessment tools, Asif et al. [8] reported alarmingly high levels of depression, anxiety and stress among Pakistani university students and Shaikh et al. [9] reported high prevalence of stress among medical students in Pakistan. All these studies collectively confirm that university students in Pakistan are experiencing psychological problems at quantifiable levels and suggest no system for detection, model for classification or any tool for addressing these problems in terms of wellness. The current paper builds directly on these findings by transforming the findings into a technology-driven actionable support system.

### *2.3 Chatbots and LLMs for Mental Wellness*

In the field of student mental health, researchers have investigated the potential of conversational AI to help alleviate student mental health issues, especially in rural developing countries. In a non-Western university setting, the low-cost solutions of a mental health chatbot using the Rasa open-source NLP system were developed and received well by the university students in Zimbabwe by Tauro and Ndlovu [18]. Their system, however, does not include any stress classification based on machine learning, and is designed for the specific context of Zimbabwe, which does not address the Pakistani cultural and linguistic specificities. One of the most appealing LLM applications for mental health, Aleem et al. [10] assessed ChatGPT's therapeutic and multicultural abilities using a series of prompting tasks and identified substantial memory, multicultural, and contextual deficits. The authors also argued that an inadequacy of current LLMs when applied to mental health services is their cultural lack of sensitivity and suggested the need for culturally adaptive AI systems that the current paper directly addresses by designing and evaluating Sukoon.

## 3. METHODOLOGY

### *3.1 Data Collection*

The data used in this study is obtained from a validated student stress survey that has been released from a public repository with 1100 college students responses across 20 features categorized into five scientifically established dimensions: psychological factors (anxiety level, self-esteem, mental health history, depression), physiological factors (headache, blood pressure, sleep quality, breathing problems), environmental factors (noise level, living conditions, safety, basic needs), academic factors (academic performance, study load, teacher-student relationship, future career concerns), and social factors (social support, peer pressure, extracurricular activities, bullying) [15]. This multi-dimensional tool was chosen because it is able to cover a wider spectrum of stressors hypothesised to impact Pakistani university students, especially the academic and relational stressors, than any single scale, such as PSS-10 [1]. Due to the time limitation for primary data collection, this study uses an incremental dataset approach, having developed and verified the whole system pipeline with this publicly available dataset. In the future, locally collected data from Ghazi University students will be included as described in section 7. The source data was de-identified and informed consent was obtained from all respondents.

### 3.2 Data Preprocessing

The output of the data collection process is subjected to a well-structured data preprocessing process before model training to ensure the quality and compatibility of the data with the machine learning algorithms used, which will directly affect the performance of prediction [12]. The first step in the preprocessing pipeline was a missing value check, which showed that the dataset had no missing values in any of the 21 columns, and thus there was no need to impute missing values. Data was then split into three segments: 70% for training, 15% for validation purposes during model tuning, and 15% for final testing, while ensuring that the distribution of classes is maintained throughout the splits. Finally, all features were normalised to zero mean and unit variance using StandardScaler, which was only fitted on the training data and applied to the validation and test sets to avoid data leakage [12].

*Fig. 1. Distribution of the target variable (stress_level) across three classes in the training dataset (n=1,100)*

### 3.3 Machine Learning Model

The Random forest algorithm is the leading classification model used in this study for the prediction of stress level. Random Forest was chosen because it has performed well on tabular survey data, has multiple decision trees that are aggregated to avoid overfitting and can produce feature importance scores that indicate which survey items are the strongest predictors of stress [12, 13]. The model takes 20 scores for the features as input and generates the probabilities for each stress level (low, moderate, and high) using their respective threshold values. Random Forest performance is then compared to Support Vector Machine (SVC) classification on the same train/test data splits to provide a strong evaluation of model effectiveness.

### *3.4 Chatbot Architecture*

Once the stress classification output that the machine learning model generates, the chatbot module will map the stress level to one of three pre-defined tiers of responses based on stress levels (low, moderate, high). The logic and sequence of how the response is generated follows the Stepped Care Model [14] which is a well recognised framework for a clinical approach of escalating intensity, tone and type of support in accordance with the level of stress detected; from a warm and encouraging approach at low stress to a grounding, calm, non-judgemental support at high stress. Cultural specificity is woven in through a well-designed system prompt that guides the underlying language model to reply in a manner consistent with Pakistani social and cultural norms including the use of Urdu and Roman Urdu expressions, as appropriate, and sensitivity to stressors unique to Pakistani University students, including family expectations, financial stress, and hierarchical teacher-student relationships [6]. This design makes sure that every interaction with the chatbot is not only appropriate for the level of stress but also culturally relevant, overcoming the main challenge that most other mental health chatbot systems face, which is the lack of cultural relevance.

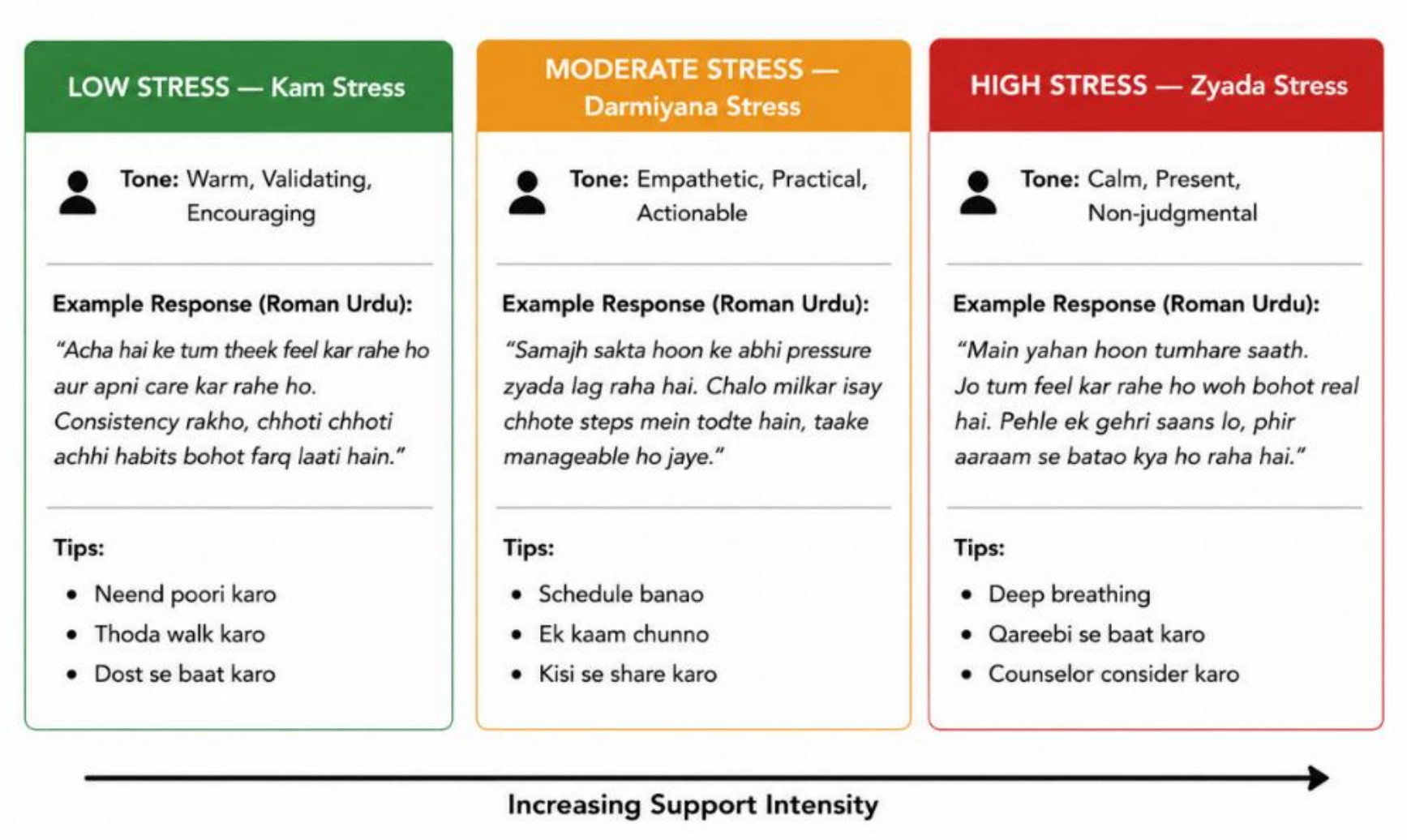

*Fig. 2. Stepped Care Model response architecture employed by Sukoon, showing how chatbot tone, language, and support intensity escalate proportionally with detected stress severity across three tiers.*

### *3.5 System Pipeline*

The proposed system starts from the student filling out a questionnaire of 20 questions in the Sukoon web application to test their stress level, followed by the system providing them with a score based on the questionnaire and recommending coping strategies. Responses go through the preprocessing pipeline described in Section 3.2, where features are normalised, and then classified into stress levels by the trained Random Forest model. The classification result, a confidence score, and stress level metadata are sent back to the front end, which then switches to the chatbot interface and automatically serves an opening wellness message tailored to the level of stress detected. Then, the GLM-4.5-Air language model processes the rest of the conversation turns through its OpenRouter API, and the entire conversation context and culturally relevant system prompt is passed in on every turn to ensure the conversation stays in context throughout the session. The entire interaction from self-report to personalised cultural wellness support is done in one seamless session. It was widely used for deploying ML models in research prototypes due to its lightweight architecture and ease of compatibility with Python-based machine learning pipelines [19], and the Flask framework was chosen.

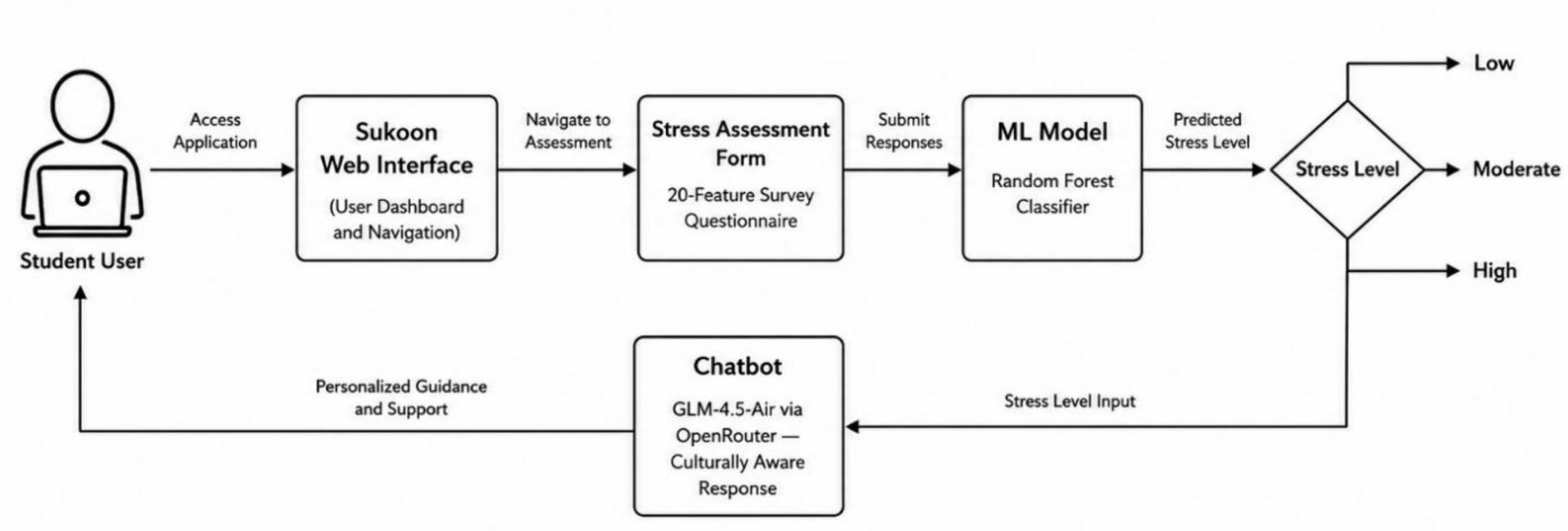

*Fig. 3. End-to-end system architecture of Sukoon, illustrating the complete pipeline from student self-report through Random Forest stress classification to culturally aware LLM-powered chatbot response delivery.*

### *3.6 Ethical Considerations*

The study was performed following the basic ethical guidelines for human-centered computing and mental health studies. The database used for model building was obtained from a public repository in which all of the responses from the participants had been anonymised and consent had been obtained by the original collectors, so that no direct ethical risk arises from the use of this data in this study. The authors acknowledge, however, that future data collection from university students in Pakistan (as is scheduled in the next stages of this research) will need to be ethically protected. All future participants will be made aware that they are being asked to participate in an activity that is voluntary, that no information of any kind will be collected, retained or used for purposes other than academic research, and that all responses will remain confidential. Consent statements will be presented at the beginning of the data collection instrument prior to any survey items. As for the Sukoon system, it is not intended to claim the chatbot as a clinical diagnostic or therapy tool. When high distress is detected, the system prompts the underlying language model to direct students to professional counselling services, and does not provide mental health services as a replacement for trained mental health providers. Some special attention has been paid in designing responses for high stress users to ensure that there is no language that could dismiss, stigmatise or cause harm. The authors' choices in design illustrate a dedication to responsible AI development in the health-adjacent space.

## 4. IMPLEMENTATION

This system was developed in Python 3.12 with the machine learning library scikit-learn [12] for the machine learning elements and the web application framework Flask. Random Forest (RF) was trained with 100 estimators and the random state 42, for consistency. We have used joblib to persist and deploy our trained model and fitted StandardScaler model. The conversational chatbot layer was developed with OpenRouter AI's API and the free-access, multilingual GLM-4.5-Air open-source language model, chosen for its ability to communicate in multiple languages and its low resource requirements, thereby facilitating its deployment in academic settings like regional universities in Pakistan.

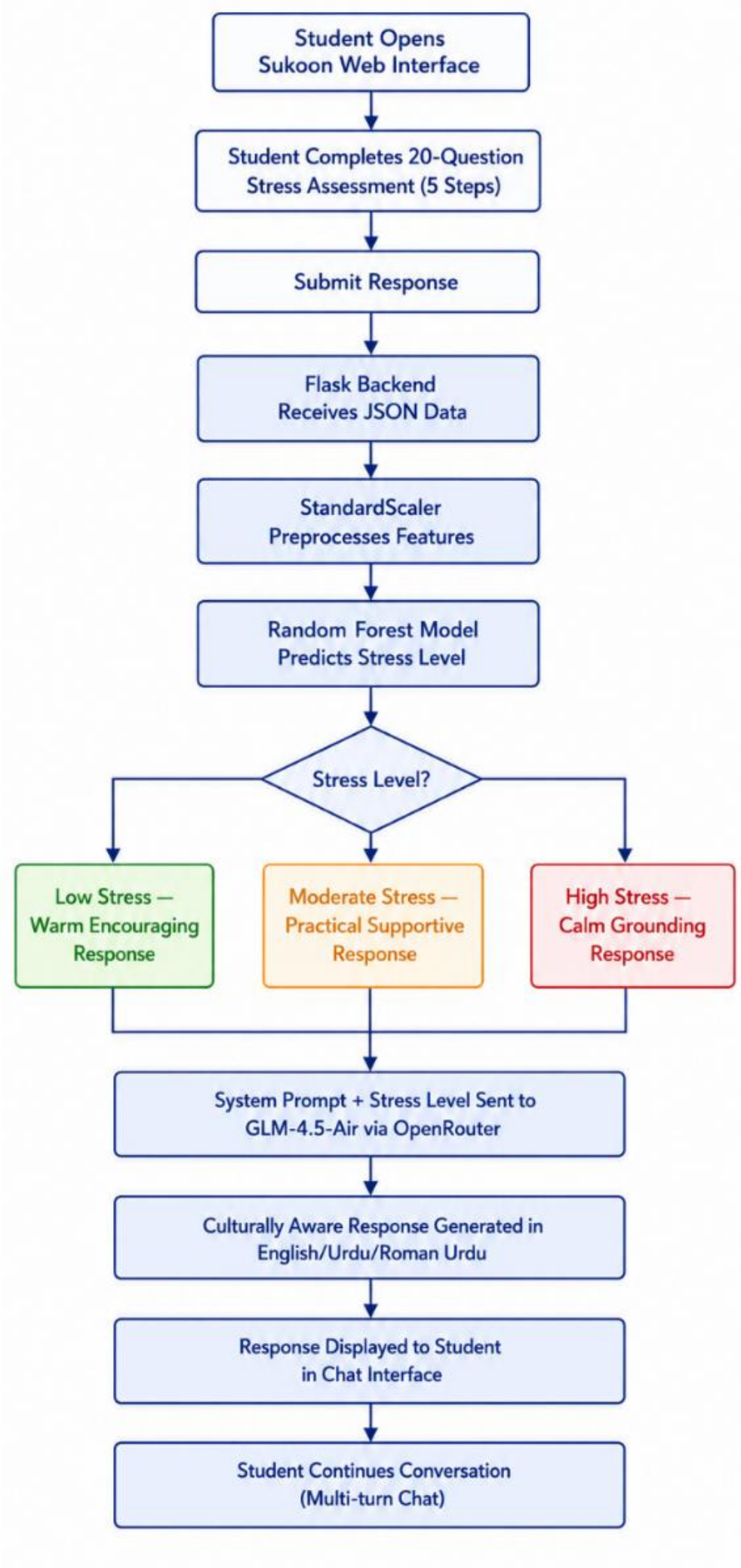

*Fig. 4. Conversation flow and operational pipeline of the Sukoon AI-based student wellness support system.*

The user interaction flow begins with the student completing a 20-question bilingual assessment presented in five thematic steps through a calm, visually accessible interface with parallel English and Urdu question labels. Upon submission, responses are transmitted to the Flask backend, preprocessed using the fitted scaler, and classified by the Random Forest model. The stress level result, confidence score, and level metadata are returned to the frontend, which transitions automatically to the chatbot interface and displays a culturally appropriate opening message. Subsequent conversation turns are powered by the LLM with a Pakistani-context system prompt, maintaining full conversation history across turns for coherent multi-turn dialogue.

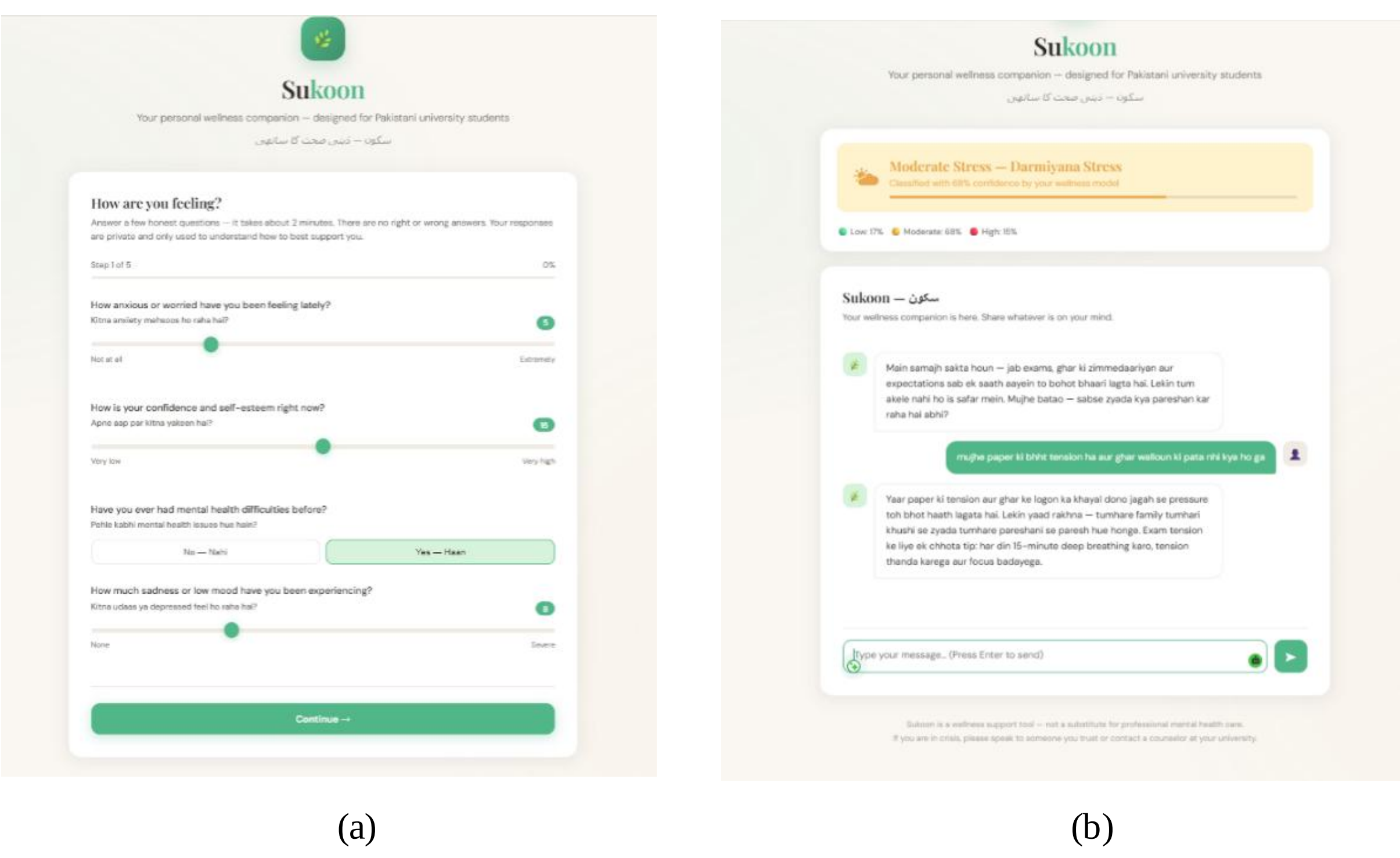

(a) (b)

*Fig. 5. Sukoon web interface: (a) bilingual stress assessment form presenting 20 survey questions across five thematic steps, and (b) culturally aware chatbot response following ML-based stress level classification.*

## 5. RESULTS AND DISCUSSION

### *5.1 Classification Results*

The proposed Random Forest classifier obtained an accuracy of 89.09% and an F1-score of 0.89 on the held-out test set with macro-average over all three stress categories. The performance of the model could be further evaluated by its performance at the class level, and the results for all three stress levels were relatively good and stable, with a F1-score of 0.87 for Low Stress, 0.92 for Moderate Stress and 0.89 for High Stress, with a relative uniformity in the results of the three classes. To make a comparison, another Support Vector Machine classifier was trained with the same data split and tested, and had an accuracy rate of 88.48% and a macro F1-score of 0.89. The proposed model is compared with related work and achieves higher accuracy than Zahra et al. [16] with 87.93% accuracy on Random Forest using a similar student stress dataset where the dataset was not as large and balanced.

*Table 1. Model Performance Comparison on Test Set*

| Metric | RF Low | RF Mod | RF High | SVM Overall | Zahra et al. [16] |
|---|---|---|---|---|---|
| Overall Accuracy | 89.09% | 89.09% | 89.09% | 88.48% | 87.93% |
| Precision | 0.94 | 0.89 | 0.85 | 0.89 | — |
| Recall | 0.80 | 0.94 | 0.93 | 0.88 | — |
| F1-Score | 0.87 | 0.92 | 0.89 | 0.89 | — |

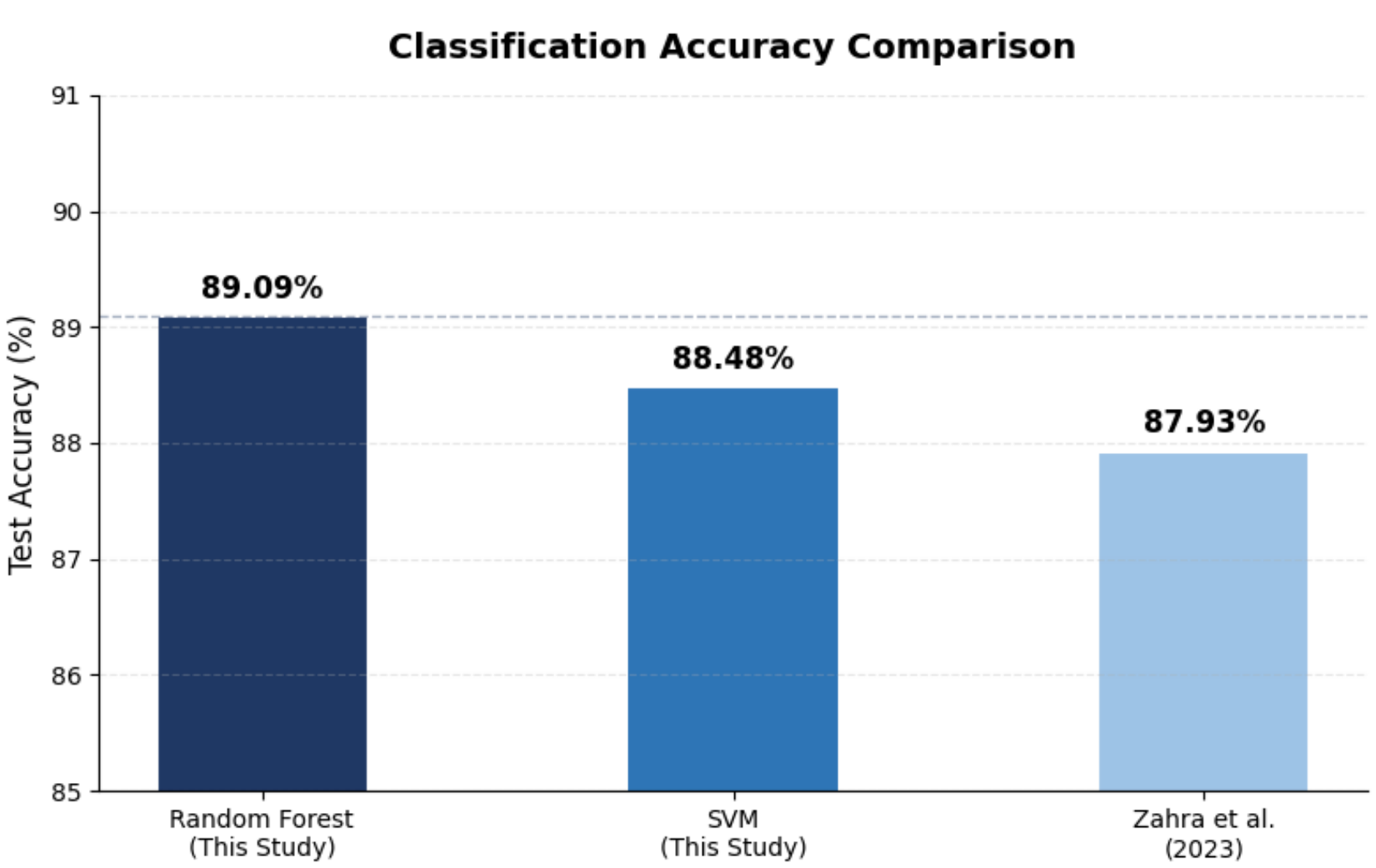

*Fig. 6. Test set accuracy comparison between Random Forest (89.09%), SVM (88.48%), and the benchmark result reported by Zahra et al. [16] (87.93%).*

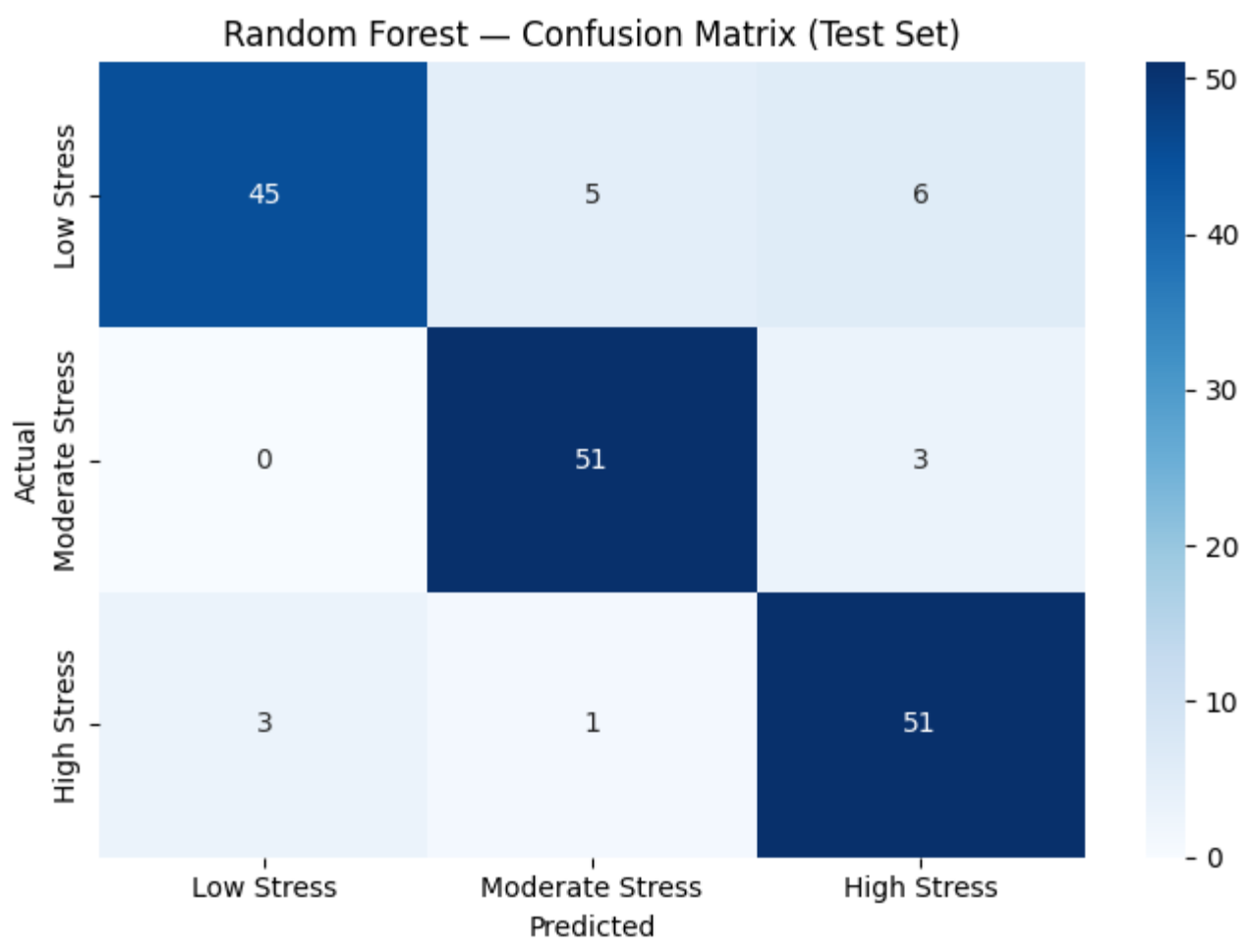

*Fig. 7. Confusion matrix of the Random Forest classifier on the held-out test set (n=165), showing predicted versus actual stress level classifications across low, moderate, and high severity categories.*

### *5.2 Feature Importance Analysis*

The feature importance analysis showed that blood pressure was the most important feature that had the most influence on the decisions for classification of student stress, accounting for 15.6% of the model's importance. This is in line with previous studies showing a strong bi-directional association between psychological stress and blood pressure increase among college students, during exam period, for example [20, 21]. The second most important predictor (10.0%) was teacher-student relationship, which was a culturally significant result described in Section 5.4. The third finding was sleep quality (9.3%) which is supported by a large body of literature documenting that poor sleep quality is a predictor and result of high stress levels in university students [22]. Notably, anxiety level, which one might intuitively expect to be the major predictor, came ninth, with only 4.8% of students reporting this. This suggests that student stress is a multi-dimensional phenomenon, rather than a phenomenon driven by a single psychological indicator, but rather is influenced by physiological, relational, and environmental factors all at once. The fifth highest was social support (7.6%), which is similar to studies identifying perceived social support as a protective factor against academic stress and depression in university students [5].

***Table 2. Top 10 Feature Importances — Random Forest Classifier***

| Rank | Feature | Importance Score |
|---|---|---|
| 1 | blood_pressure | 0.1561 |
| 2 | teacher_student_relationship | 0.1002 |
| 3 | sleep_quality | 0.0932 |
| 4 | depression | 0.0833 |
| 5 | social_support | 0.0757 |
| 6 | self_esteem | 0.0732 |
| 7 | bullying | 0.0512 |
| 8 | safety | 0.0482 |
| 9 | anxiety_level | 0.0481 |
| 10 | headache | 0.0449 |

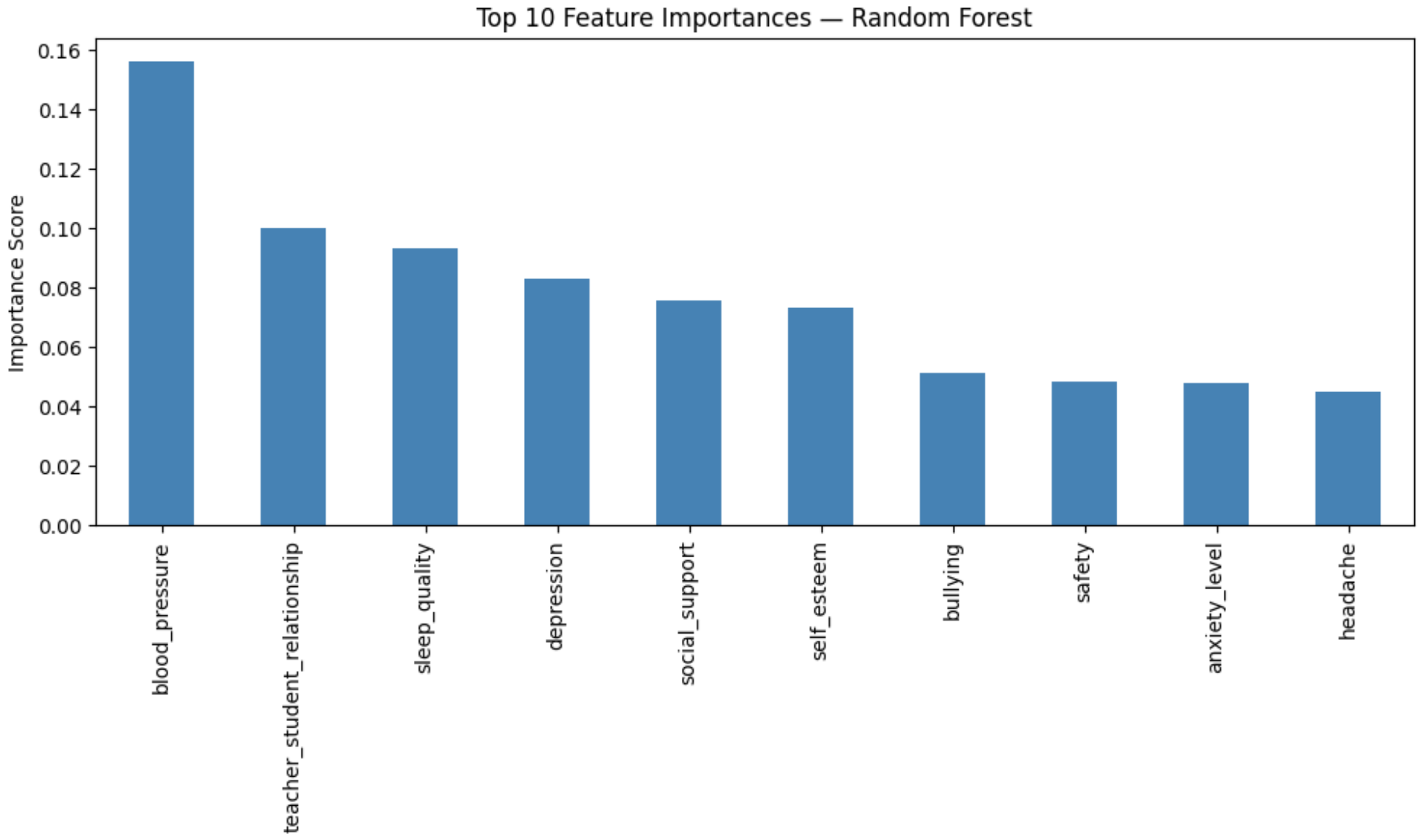

*Fig. 8. Top 10 feature importance scores generated by the Random Forest classifier, identifying the strongest predictors of student stress level from the 20-feature survey dataset.*

### *5.3 Discussion of Model Performance*

The Random Forest model achieved a higher accuracy on the test set compared to SVM because of its ensemble structure that combines predictions from multiple decision trees, resulting in more reliable and consistent predictions on tabular survey data [12, 13]. We consider an accuracy of 89.09% to be adequate for deployment in a student wellness application, in addition to the fact that it is equipped with an open source LLM as a conversational layer that delivers intelligent and adaptive support beyond the classification result. The combination of the ML classification backend and the LLM-based frontend forms a hybrid system where the decision regarding the tone and level of support is informed by the model's confidence, whereas the conversational AI helps generate responses that are contextually relevant, linguistically appropriate, and culturally relevant to the Pakistani students using the service, provided they are proficient in English, Urdu, or Roman Urdu [10].

### *5.4 Discussion of Cultural Findings*

Teacher-student relationship is the second most predictive stress feature which has a significant cultural implication for Pakistani university students [6]. The educational environment in Pakistan is much more hierarchical, with a lack of academic support for students, less encouragement, and a high degree of power imbalance; these stressors are not accounted for or addressed in mental health tools based in Western educational settings [2, 3]. The result empirically proposes — as a hypothesis — that culturally specific relational factors may be measurable and explainable as a factor of student stress which is not represented in the global datasets. Concerning classification errors, the confusion matrix analysis showed that the main misclassification pattern of the model was that Low Stress students were classified as Moderate Stress students. This is a safe and acceptable error in a wellness deployment context: A student who is erroneously placed in a higher stress category gets extra support; in the deployment context, it is not harmful. The mean of the errors is the one that was truly dangerous: treating a high stress student as a low stress student; it was uncommon, and the system is correctly biased in the clinically appropriate direction [14].

### *5.5 Chatbot Response Evaluation*

To check for the functional coherence and cultural appropriateness of Sukoon's chatbot responses before wider deployment, a preliminary qualitative evaluation of the chatbot responses was carried out in addition to evaluating the machine learning classification model. The system was validated in all three stress level categories (low, moderate, and high) using simulated user inputs that encompass some of the most frequent concerns of Pakistani University students: exam-related stress, family stress, financial stress, and teacher-related stress. In all cases, the system was able to identify the correct stress tier from the classification output and respond to it in an appropriate way, within a Stepped Care Model framework described in Section 3.4.

Informal usability testing showed that the process of switching from the assessment form to the chatbot interface was smooth and the bilingual English and Roman Urdu answers were found to be natural and culturally relevant by the tester. The system could preserve coherent, multi-turn conversation history over long conversations without losing the sense of the context. The authors, however, recognise that this is a preliminary functional assessment and not a formal user study. A comprehensive evaluation with a representative sample of Pakistani University students, measuring the perceived cultural appropriateness, emotional safety, response relevance, and user satisfaction will be an important focus of future work, and will be carried out in conjunction with the primary data collection period outlined in Section 7.

## 6. CONCLUSION

In this paper we introduced a culturally sensitive stress detection and wellness support system named Sukoon, which also has an artificial intelligence based feature [2, 3, 8]. A validated student stress dataset of 1,100 responses was used to develop and evaluate a Random Forest classification model, with test accuracy of 89.09% and a macro F1-score of 0.89 for three levels of student stress. The classification output was fused with an open-source large language model (LLM) which provides personalised wellness support in a culturally nuanced tone and language that aligns with Pakistani cultural values, making Sukoon the first system of its kind to integrate ML-based stress classification with culturally adapted AI dialogue for the Pakistani context.

Students of universities in Pakistan are a group of students who have a very complex set of stressors that are not addressed by mental health tools designed in western countries [10, 11] and therefore, particularly important in the context of the proposed system. The empirical result of feature importance analysis indicated that teacher-student relationship is one of the most important factors in determining student stress, which is a culturally significant result and also supports the need to prepare mental health systems that are region specific [6]. Sukoon's design is built on an awareness of the environmental and relational stresses unique to the university context in Pakistan and responding to them with a design appropriate to this space and population that has been largely unaddressed by digital mental health tools. Directions for further research and development of the system are detailed in Section 7.

## 7. FUTURE WORK & LIMITATIONS

### *7.1 Future Work*

Future work will focus on collecting primary data from Pakistani university students using an Urdu-translated and clinically validated DASS-21 instrument to improve the cultural relevance of the classification model [1]. The collected data will be used to retrain and evaluate the proposed system using locally representative stress patterns, particularly among freshman students transitioning from FSc to university life. In addition, a formal user evaluation study will be conducted to assess the chatbot's cultural appropriateness, emotional safety, response relevance, and overall user satisfaction. Future versions of Sukoon may also incorporate longitudinal journaling features, mobile application support, and integration pathways with university counselling services to enhance accessibility and long-term student wellness support.

### *7.2 Limitations*

There were several drawbacks to this study. First, the training data is not representative of the Pakistani students, and also might not be reflective of the local cultural and socioeconomic stressors, therefore the feature importance results are preliminary conclusions that need to be confirmed with data collected locally [15]. Second, there is a lack of formal evaluation of the chatbot responses using user studies with Pakistani university students. Third, the system is currently in English and Urdu expressions are prompted in the system, not by a fully bilingual NLP pipeline. Fourth, the evaluation of the classifier was performed on one stratified split; k-fold cross validation is used in future to get more solid performance estimates with confidence intervals [12]. The scope of future work is outlined in Section 7.1 and is limited by these constraints.